\documentclass[]{interact}

\usepackage{epstopdf}
\usepackage[caption=false]{subfig}

\usepackage[numbers,sort&compress]{natbib}

\usepackage{booktabs}
\usepackage{todonotes,soul}
\usepackage{amsmath}
\usepackage{enumitem}

\usepackage{graphicx}
\usepackage{mathtools} 

\usepackage{babel}
\usepackage{algorithm}
\usepackage{algpseudocode}

\bibpunct[, ]{[}{]}{,}{n}{,}{,}
\renewcommand\bibfont{\fontsize{10}{12}\selectfont}
\makeatletter
\def\NAT@def@citea{\def\@citea{\NAT@separator}}
\makeatother

\theoremstyle{plain}

\theoremstyle{definition}

\theoremstyle{remark}

\begin{document}

\articletype{ARTICLE TEMPLATE}

\title{Automatic Detection of Complex Quotation Patterns in Aggadic Literature}


\author{
\name{Hadar Miller\textsuperscript{a}\thanks{CONTACT Hadar Miller. Email: hmille02@campus.haifa.ac.il}, Tsvi Kuflik\textsuperscript{a} and Moshe Lavee\textsuperscript{a}}
\affil{\textsuperscript{a}University of Haifa, Haifa, Israel;}
}

\maketitle

\begin{abstract}
This paper presents ACT (Allocate Connections between Texts), a novel three-stage algorithm for the automatic detection of biblical quotations in Rabbinic literature. Unlike existing text reuse frameworks that struggle with short, paraphrased, or structurally embedded quotations, ACT combines a morphology-aware alignment algorithm with a context-sensitive enrichment stage that identifies complex citation patterns such as “Wave” and “Echo” quotations.

Our approach was evaluated against leading systems—including Dicta, Passim, Text-Matcher—as well as human-annotated critical editions. We further assessed three ACT configurations to isolate the contribution of each component. Results demonstrate that the full ACT pipeline (ACT-QE) outperforms all baselines, achieving an $F_1$ score of 0.91, with superior Recall (0.89) and Precision (0.94). Notably, ACT-2, which lacks stylistic enrichment, achieves higher Recall (0.90) but suffers in Precision, while ACT-3, using longer n-grams, offers a tradeoff between coverage and specificity.

In addition to improving quotation detection, ACT’s ability to classify stylistic patterns across corpora opens new avenues for genre classification and intertextual analysis. This work contributes to digital humanities and computational philology by addressing the methodological gap between exhaustive machine-based detection and human editorial judgment. ACT lays a foundation for broader applications in historical textual analysis, especially in morphologically rich and citation-dense traditions like Aggadic literature.

\end{abstract}


\section{Introduction}
\label{introduction}
Rabbinic Literature of Late Antiquity, and especially the Midrash, is centered on the Hebrew Bible. The texts playfully interpret the biblical texts, develop them, and apply them to various legal, moral, religious, and literary goals. Among the key features of Midrashic hermeneutics is the constant move from dwelling on very small textual units (even single words or syllables) to reading the entire corpus as one deeply interrelated entity.

To understand the connections between Midrashic texts and the Bible, it is essential to detect their intertextual links. Automatic text reuse detection frameworks refrained from detecting single-word quotations, as such quotations are inherently challenging to identify and may correspond to multiple occurrences of the same word across different contexts \cite{shmidman2022automatic}. However, Rabbinic literature frequently uses quotations of a single word. For example, in the fifth-century midrash Genesis Rabbah 1:7, in the phrase "as it is said, 'Bereshit', etc.' (Gen. Rab. 1.7). The word "Bereshit" refers to Gen. 1:1. In addition, the author employs specific terminological markers such as: "as it said" and "etc." to signal the beginning and end of a quotation. Despite this clear marker, automatically detecting the quotation remains challenging due to the word's multiple occurrences in the Bible (e.g., Jer. 26:1; 27:1; 49:34; Isa. 28:1). In such cases, the surrounding context plays a crucial role in clarifying the intended reference. In such cases, contextual analysis is essential for disambiguating the intended source. In the example above, the homily centers on the beginning of the Book of Genesis and includes multiple references to Genesis, making it evident to the human reader that Bereshit refers to Genesis 1:1, rather than to other biblical instances.

The necessity of detecting single-word quotations arises from the broader objective of identifying quotation patterns within rabbinic texts. Midrashic literature commonly employs recurring quotation patterns that involve dissecting verses into individual words and interchanging citations from multiple verses. This characteristic necessitates a dual-stage detection process: first, identifying individual quotations, including single-word quotations, and second, recognizing the overarching quotation patterns they form.

Text reuse detection frameworks such as \cite{smith2014detecting}, \cite{buchler2016tracer}, and \cite{shmidman2022automatic} have proven effective for detecting Biblical quotations in ancient languages, including Hebrew. However, they are limited in detecting very short quotations. They all rely on n-grams comparison with no contextual filtering. This results in relatively high false-negative rates when larger n-grams are used and high false-positive rates when small n-grams are used.

This study proposes a pipeline that first segments the text into word-level n-grams using shingling, as described in \cite{buchler2016tracer}. Next, intersection n-grams with the biblical text are identified, following the approach of \cite{smith2014detecting} while replacing the alignment algorithm with \cite{miller2025text}. An n-gram size of one enables the detection of single-word quotations.

Quotation candidates are stored in a graph structure that preserves the sequential order of quotations within the text. These candidates are then scored based on three factors: the alignment score, the rarity of word combinations—following the methodology of \cite{shmidman2022automatic}—and the quotation's context. The latter factor also enriches each quotation by identifying its corresponding quotation pattern. Finally, a graph pruning step removes candidates with low scores.

Using a ground-truth dataset manually annotated by experts, we demonstrate that our approach improves the $F_1$ score by 12\% compared to the state-of-the-art framework \cite{shmidman2022automatic}.

Two primary research goals in studying rabbinic texts could benefit from such a detection process. First, parallel rabbinic sources may share identical quotation patterns despite significant differences in phrasing and even language (e.g., shifts between Hebrew and Aramaic). In such cases, quotation pattern classification can facilitate the identification of paraphrased text reuse. Second, midrashic works can often be characterized by the specific citation patterns they employ, which may serve as indicators of a text's date or subgenre. Accordingly, identifying single-word quotations, followed by detecting broader quotation patterns, can contribute to distinguishing between literary genres and uncovering different compositional layers within a single work.

\section{Related Work}
\label{Related Work}

The automatic detection of text reuse—especially within historical and religious texts—has been widely explored, given its significance in uncovering intertextual relationships and tracing cultural transmission. One of the earliest frameworks, proposed by \cite{kolak2008generating}, approaches quotation detection by segmenting the corpus into n-grams and identifying overlapping sequences between a target text and the corpus. Text reuse is inferred when the length of these overlapping sequences exceeds a predefined threshold. However, this method proves less effective when identifying quotations that include even minor orthographic variations. This limitation is particularly relevant in the context of Hebrew, a language that lacks standardized orthography \cite{siegal2018reconstruction}.

\cite{smith2014detecting} introduced a more robust framework consisting of three sequential steps: (1) character-level n-gram matching to identify document pairs with a high likelihood of local text reuse, (2) local text alignment for each identified pair, and (3) similarity measurement to confirm textual relationships between documents or their respective sections formally. Their implementation employed the algorithm developed by \cite{smith1981identification} in its simplest form for local alignment. While this approach effectively detects substantial textual overlap, it struggles to identify shorter quotations \cite{smith2014detecting}. Similarly, Text-matcher \cite{Reeve2020} is another straightforward string-matching algorithm for detecting text reuse, but unlike \cite{smith2014detecting}, it operates by matching word-level n-grams. The effectiveness of this method diminishes when quotations contain spelling differences or lexical substitutions, such as the use of synonyms.

\cite{kokkinakis2015detecting} employed the TextPAIR framework introduced by \cite{olsen2011something} to detect biblical quotations within a corpus of 19th-century Swedish prose fiction. More recently, \cite{gladstone2023intertextual} utilized the same framework to develop an intertextual hub for studies of 18th-century French literature. TextPAIR identifies text reuse by matching n-grams between an analyzed text and a corpus, subsequently applying a sequence alignment algorithm to infer and visualize reuse instances. Similar to the method of \cite{smith2014detecting}, TextPAIR is "largely effective in identifying similar passages with extensive variations" \cite{olsen2011something}; however, it remains less effective for detecting shorter quotations.

\cite{buchler2016tracer} advanced the field by introducing Tracer, a tool specifically designed to detect text reuse in ancient languages. Tracer employs a multi-stage process, beginning with preprocessing steps that normalize tokens using researcher-defined algorithms, such as diacritic removal and synonym replacement. Although the default configuration does not include preprocessing algorithms tailored for Hebrew, the framework is extensible and allows for language-specific adaptations. Tracer offers two principal methods for detecting text reuse based on term features: (1) a word-based feature, which is suitable for identifying paraphrastic reuse, and (2) an n-gram-based feature, which is primarily effective for detecting verbatim reuse \cite{franzini2017user}. However, while the n-gram-based method performs well in identifying longer segments of verbatim reuse, it is too rudimentary for capturing very short instances of textual similarity.
 
Shmidman \cite{shmidman2022automatic} proposed a state-of-the-art algorithm for automatically detecting biblical citations in Hebrew. This algorithm preprocesses biblical texts into skip n-grams ranging in size from two to five tokens and generates base forms for each skip n-gram using an existing treebank of the Hebrew Bible. Similar preprocessing steps are applied to the target text, after which overlapping skip n-grams are identified. Text reuse is inferred when matched n-grams yield a unique form corresponding to a limited set of biblical verses. Despite its innovative approach, this algorithm, by design, cannot detect single-word quotations \cite{shmidman2022automatic}.

\section{Biblical Quotation Styles}
\label{Biblical Quotations Typography}

Biblical quotations may consist of a few words copied verbatim. However, such quotations are frequently modified in rabbinic sources through slight alterations or paraphrasing. These paraphrases can often be categorized into distinct types based on identifiable patterns. This study aims to define several major quotation styles to facilitate their automatic detection within texts.

Building on the 25 classes of quasi-paraphrases identified by \cite{bhagat2013paraphrase}—including synonym substitution, insertion of external knowledge, repetition, and other transformations—we apply this typology specifically to classify paraphrastic elements within biblical quotations. Through empirical analysis, we identified four overarching patterns of quotation in Rabbinic literature: Simple, Wave, Echo, and Compound (detailed in later sections). The Simple and Wave patterns encompass paraphrastic quotations, where Bhagat’s classification helps us describe the types of transformations involved. The Echo and Compound forms represent combinations or recursive applications of the first two types. Thus, rather than offering a comprehensive mapping of all 25 paraphrase classes, we abstracted them into four broader quotation styles tailored to the specific characteristics of biblical reuse in our corpus. We define the distinctive features of each pattern and assess how effectively existing text reuse detection frameworks capture these nuanced variations.

\begin{figure}[h!]
    \centering
    \includegraphics[width=\linewidth]{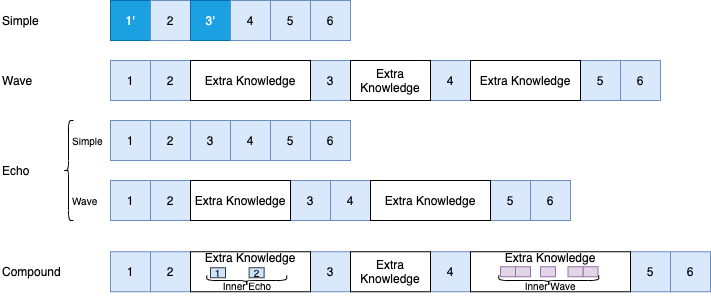}
    \caption{Structure of the Different Quotation Styles}
    \label{fig:quotation_styles}
\end{figure}

Figure \ref{fig:quotation_styles} illustrates the high-level structure of each quotation style pattern. The Simple style consists of a sequence of verbatim biblical words, though minor orthographic or synonymic modifications may occur, as seen in the first and third words of the example. The Wave style permits the insertion of external knowledge between segments of the biblical quotation. The Echo style combines two patterns—Simple and Wave—referencing the same biblical verse. Finally, the Compound style extends the Wave pattern by incorporating additional biblical quotations, of any style, within its interleaved external segments.

The following sections provide a detailed discussion of each quotation style.

\subsection{"Simple" Quotation Style}
The Simple quotation style typically involves short quotations with minimal modifications, such as orthographic variations or synonym substitutions. \cite{bhagat2013paraphrase} classifies these as Class 1 synonym substitutions. This style is generally amenable to detection by existing text reuse frameworks, as demonstrated by \cite{smith2014detecting}, \cite{buchler2016tracer}, and \cite{shmidman2022automatic}. However, when quotations are extremely short, their identification increasingly relies on contextual cues within the surrounding text to establish intertextual relationships. Nevertheless, many frameworks—including \cite{shmidman2022automatic}—explicitly avoid detecting single-word quotations. Those that do permit single-word n-grams often exhibit high false-positive rates, largely due to inadequate context-sensitive filtering.

\subsection{"Wave" Quotations Styles}

The Wave quotation style paraphrases biblical texts by inserting additional information between the words of the verse. \cite{bhagat2013paraphrase} classifies this pattern as “Class 25 – Extra Knowledge.” In this style, a biblical verse is segmented into multiple parts, each interspersed with commentary or elaboration intended to provide additional context for the reader. These segments may sometimes be as short as a single word.

\begin{figure}[h!]
    \centering
    \includegraphics[width=\linewidth]{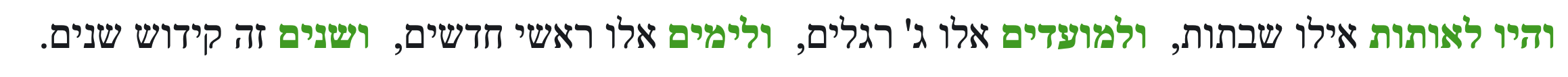}
    \caption{"Wave" Style Quotation, Albek edition of Bereshit Rabba, 3 7}
    \label{fig:quotation_wave}
\end{figure}

Figure \ref{fig:quotation_wave} illustrates this style using an excerpt from the Albek edition of Bereshit Rabba 3:7, in which a biblical verse (Genesis 1:5) is interwoven with interpretative commentary. This style requires alternating between quoted and explanatory text, resulting in a wave-like textual pattern. Due to the fragmentation of the quotation, standard n-gram-based methods employed in text reuse detection frameworks face considerable difficulty in accurately identifying this style.

\subsection{"Echo" Quotations Styles}
The Echo quotation style consists of a sequence of two quotations from the same biblical verse. The first appears in the Simple style, while the second adopts the Wave style. The ability of existing text reuse detection frameworks to identify Echo quotations is limited by the challenges involved in individually detecting each component style.

\begin{figure}[h!]
    \centering
    \includegraphics[width=\linewidth]{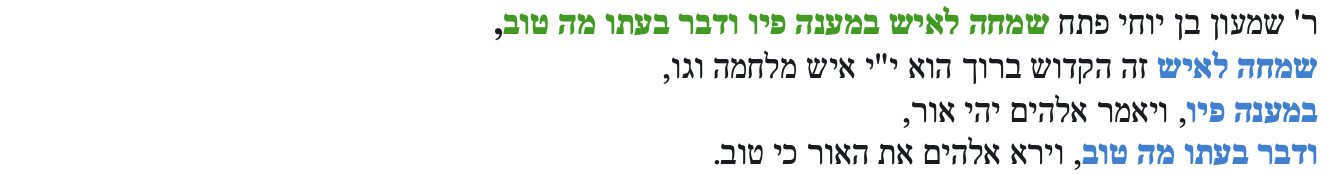}
    \caption{"Echo" Style Quotation,  Albek edition of Bereshit Rabba,  3,3. The phenomenon head is marked in green, while the phenomenon tail is marked in blue.}
    \label{fig:quotation_repetition}
\end{figure}

Figure \ref{fig:quotation_repetition} illustrates this style using an excerpt from the Albek edition of Bereshit Rabba 3:3. In this example, the initial quotation from Proverbs 15:23—"Simḥa le'ish be-ma'aneh piv u-davar be-ito mah tov" (highlighted in green)—is presented in the Simple style. This is followed by a Wave style quotation of the same verse, such as "Simḥa le'ish" (highlighted in blue), interwoven with elaborations from other biblical sources. Existing frameworks often struggle to detect these brief repetitions, as they typically fail to account for the broader intertextual context in which the quotation occurs.

\subsection{"Compound" Quotations Styles}
The Compound quotation style builds upon the Wave style by embedding additional biblical quotations within the extra-knowledge segments of the wave structure. Each of these embedded quotations may follow any of the defined quotation styles.

This intricate literary structure increases the distance between the components of the original Wave quotation, thereby complicating their automatic identification by text reuse detection frameworks.

\begin{figure}[!htb]
    \centering
    \includegraphics[width=\linewidth]{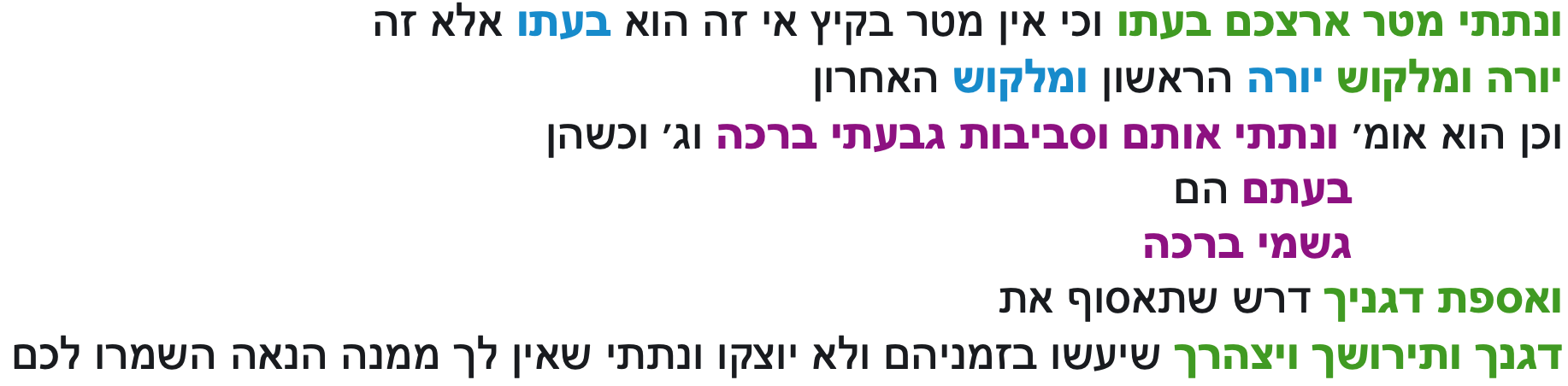}
    \caption{"Mixture of different quotation styles, Vachman edition of New Midrash, Vaetchanan 6}
    
    \label{fig:quotation_mix}
\end{figure}

Figure \ref{fig:quotation_mix} presents an example from the Vachman edition of New Midrash, Vaetchanan 6, which exhibits three layers of quotations. The outermost layer features a quotation from Deuteronomy 11:14 (highlighted in green) that follows the Wave style, fragmenting the biblical verse into four parts.

The extra-knowledge segment of the first Wave component contains the author's elaboration. The second extra-knowledge segment adopts the Echo quotation style (highlighted in blue) and is immediately followed by an additional biblical quotation, which follows the Wave style (highlighted in purple).

\section{Methodology}
\label{Methodology}

Our framework, ACT (Allocate Connections between Texts), comprises a three-stage pipeline, illustrated in Figure~\ref{fig:pipeline}. The first two stages follow practices common in existing text reuse frameworks, while the third stage is a novel contribution introduced in this work to handle complex biblical quotation patterns.

\textbf{Stage 1: Preprocessing.} This stage enhances recall by addressing orthographic, morphological, and semantic variation through normalization techniques. It includes the removal of diacritics, standardization of spelling variants, and tokenization.

\textbf{Stage 2: Candidate Detection.} The input text is segmented into overlapping n-grams. For each n-gram, potential quotation candidates are retrieved using a positional inverted index \cite{manning2008introduction} and aligned using a sequence alignment algorithm tailored for morphologically rich languages \cite{miller2025text}. This innovation enables greater robustness in handling Hebrew and other similarly complex languages.

\textbf{Stage 3: Quotation Enrichment and Pattern Inference.} In this novel stage, quotations are not only scored by uniqueness and contextual cues but are also enriched with metadata identifying their structural style (e.g., Wave or Echo). This stage enables the detection of quotations that are short, embedded, or paraphrased—cases typically missed by standard reuse methods.

\begin{figure}[h!]
\centering
\includegraphics[width=\linewidth]{figures/Block_diagram_of_Quotation_Detection_Pipeline.png}
\caption{Block diagram of ACT’s three-stage pipeline for detecting biblical quotations: (1) *Preprocessing* normalizes both corpus and examined texts; (2) *Candidate Detection* identifies and aligns potential quotations using a sliding window and a positional inverted index; (3) *Quotation Inference* ranks, filters, and classifies quotations into four styles. Colored arrows indicate the flow of text data (green), index lookup (blue), and quotation inference (gold).}

\label{fig:pipeline}
\end{figure}

\subsection{Preprocessing Stage}
\subsubsection{Text Normalization}
To improve alignment accuracy, ACT applies a series of preprocessing steps that normalize and clean both the input texts and the biblical corpus. These steps are designed to reduce orthographic and structural variability:

Vocalization Removal: Diacritical marks (e.g., Hebrew niqqud) are removed to reduce phonetic noise and orthographic variability.

Special Character Cleaning: Special symbols (e.g., \%, \$) and typographically inconsistent characters (e.g., varying quotation marks) are standardized to prevent alignment errors.

Matres Lectionis Removal: Vowel-indicating letters are stripped to further reduce orthographic variation, especially relevant in Hebrew.

Tokenization: Texts are segmented into tokens.

These steps are applied consistently to both the corpus and the input text, ensuring uniformity in representation.

\subsubsection{Positional Inverted Index}
Quotation detection requires a digital corpus of biblical texts. We utilized \cite{Sefaria}, which underwent the same normalization and cleaning processes described above.

From this corpus, we constructed a positional inverted index that maps each token to its exact positions across the corpus. This structure enables rapid retrieval of potential matches and significantly reduces computational overhead in the candidate detection phase.

\subsection{Candidate Detection Stage}
The input text is segmented into overlapping n-grams using a sliding window, where both the window size and stride are user-defined parameters. The window size controls the length of each n-gram, while the stride determines the degree of overlap.

Each n-gram is used to query the positional index to identify potentially matching segments in the biblical corpus. Candidate pairs are then aligned using the morphologically-aware alignment algorithm introduced in \cite{miller2025text}, which accounts for paraphrastic expressions, orthographic variation, transcriptional noise, and word order differences.

\subsection{Quotation Inference Stage}
\label{Quotation Inferupdate ence Stage}

The final stage determines whether a detected sequence constitutes a biblical quotation and, if so, infers its quotation style.

Conventional approaches evaluate each candidate pair independently, using heuristic thresholds based on match length or overlap percentage. However, this leads to high false-negative rates in cases of fragmented reuse, such as in the Compound or Wave styles, where no individual match surpasses the detection threshold.

To overcome this, we propose a probabilistic scoring method optimized for short quotations. Each detected sequence is initially classified as Simple and assigned a score derived from the log-likelihood of its constituent words. The rarity of a sequence is estimated by computing the product of the probabilities of its individual tokens across the corpus. To avoid numerical underflow, we instead sum their log-probabilities, which preserves relative rarity and allows for stable thresholding.

We propose a probabilistic scoring method optimized for short quotations. Each detected sequence is initially classified as Simple and assigned a score derived from the log-likelihood of its constituent words.
For example, consider the detected quotation "Bereshit Bara" (“In the beginning created”), aligned with Genesis 1:1. The probability of the sequence is computed as the product of the individual token probabilities across the biblical corpus:
$$P("Bereshit Bara") = P("Bereshit")XP("Bara")$$
To avoid numerical underflow, we instead sum their log-probabilities:
$$P("Bereshit Bara") = Log( P("Bereshit") ) + Log( P("Bara") )$$
This approach preserves the relative rarity of tokens while enabling stable thresholding for quotation inference. As a result, frequently quoted sequences receive higher scores, whereas improbable or coincidental co-occurrences are effectively penalized.

To account for non-contiguous quotations characteristic of Wave or Echo styles, we introduce an inference algorithm (Algorithm~\ref{alg:boost_rank}) that scans for prior quotations from the same verse and boosts their scores if intertextual patterns are observed. If multiple quotations are identified as sharing context (i.e., proximity or intervening text), their scores are aggregated, and a style label (wave or echo) is assigned accordingly.

\begin{algorithm}[H]
\caption{Boost Quotation Rank and Label Style}
\label{alg:boost_rank}
\begin{algorithmic}[1]
\Require Detected quotations $Q = {q_1, q_2, \dots, q_n}$, each with verse and text
\Ensure Updated ranks and style labels for all quotations

\For{$i = 2$ to $n$}
\State $p \Leftarrow f(q_{\text{verse}}, i)$ \Comment{Locate previous quotation of same verse}
\If{$p$ exists}
\State Boost ranks of $q_i$ and $p$ by summing individual ranks
\If{$q_i$ and $p$ share intervening text}
\State Set style label of $q_i$ and $p$ to \texttt{"echo"}
\Else
\State Set style label of $q_i$ and $p$ to \texttt{"wave"}
\EndIf
\EndIf
\EndFor

\Return Updated ranks and style labels
\end{algorithmic}
\end{algorithm}

\section{Experiments}
\label{Experiments}

To assess the effectiveness and innovations of our proposed method, we conducted an evaluation of ACT against several prominent systems for detecting biblical quotations in Hebrew texts. These include \cite{shmidman2022automatic} (referred to as Dicta), \cite{smith2014detecting} (Passim), \cite{Reeve2020} (Text-Matcher), and manually annotated quotations from critical editions.

Our evaluation focused on two key objectives: (1) comparing the overall performance of ACT with existing methods, and (2) measuring the impact of each innovation in ACT through controlled configuration experiments. This dual analysis allowed us to highlight both the relative advantages of ACT and the unique contributions of its individual components.

\subsection{ACT Configurations}
To isolate and assess the contributions of our improvements, we evaluated three configurations of the ACT algorithm:

\begin{itemize}
    \item \textbf{ACT-2:} Incorporates only Improvement A—our morphology-aware alignment algorithm—executing the first two stages of the pipeline with an n-gram size of 2.
    \item \textbf{ACT-3:} Also includes only Improvement A, but with an n-gram size of 3 to examine the effect of n-gram granularity.
    \item \textbf{ACT-QE:} Combines both Improvement A and B—i.e., the full three-stage pipeline—with an n-gram size of 1. This is the complete version of ACT.
\end{itemize}

\subsection{Baselines}
For comparison, we included the following established methods:

\begin{itemize}
    \item \textbf{Dicta} \cite{shmidman2022automatic}: A state-of-the-art system for automatic Hebrew quotation detection.
    \item \textbf{Passim} \cite{smith2014detecting}: A general-purpose alignment framework using character-level matching.
    \item \textbf{Text-Matcher} \cite{Reeve2020}: An n-gram-based reuse detection algorithm.
    \item \textbf{Critical Editions:} Manual annotations by expert editors in scholarly editions, serving as a human-curated gold standard.
\end{itemize}

\subsection{Dataset}
\label{Dataset}
To establish a reliable benchmark for evaluation, we constructed a manually annotated dataset with the assistance of domain experts. The dataset includes three compositions: the Albek edition of Bereshit Rabba \cite{theodor1965bereshit}, the Milikovski edition of Vayikra Rabba \cite{vayikrarabba}, and the Wachman edition of New Midrash \cite{vachman2013midrash}.

Annotators were instructed to label the exact boundaries of each quotation, identify the associated biblical verse, and classify the quotation style. In cases of Wave style, each fragment was annotated as an independent quotation, but all fragments were linked to the same biblical verse. In the Echo style, both the initial Simple quotation and its repeated fragments (Wave) were annotated separately.

Each quotation was represented as a five-tuple, adapted from \cite{alvi2021paraphrase}: $s = (s_{start}, s_{size}, b_{verse}, b_{start}, b_{size})$ 

Where:
\begin{itemize}
\item $s_{\text{start}}$: Starting index of the quotation in the source text
\item $s_{\text{size}}$: Number of words in the quoted segment
\item $b_{\text{verse}}$: Biblical verse being quoted
\item $b_{\text{start}}$: Starting index within the biblical verse
\item $b_{\text{size}}$: Number of words from the verse being quoted
\end{itemize}

This detailed tagging scheme provides a robust ground truth (GT) for evaluating the accuracy and stylistic sensitivity of each method.

\subsection{Evaluation Metrics}
Each detected quotation was evaluated against a manually annotated ground truth using standard information retrieval metrics: Precision, Recall, and $F_1$ score. Quotations were matched using a five-tuple representation as described in Section~\ref{Dataset}.

\subsection{Style Usage in Different Literary Works}

We analyzed the distribution of quotation styles across the three literary works. Each exhibited distinct stylistic tendencies. While all texts included a substantial proportion of Simple quotations (approximately 50–60\%), the Vayikra Rabba corpus was marked by a high prevalence of the Echo style (41\%) and minimal use of the Wave style (2\%). In contrast, both Bereshit Rabba and New Midrash showed substantial use of Wave-style quotations.

Table~\ref{tab:quotation_style_dist} summarizes these distributions. These stylistic patterns point toward a promising direction for future research—namely, the classification of literary works or redactional layers based on their stylistic profiles of biblical quotation usage.

\begin{table}[h!]
  \begin{tabular}{lccl}
    \toprule
    All & Bereshit Rabba & Vayikra Rabba & New Midrash \\
    \midrule
    Simple   & 55\%          & 48\%          & 60\% \\
    Echo     & 28\%          & 25\%          & 12\% \\
    Wave     & 17\%          & 27\%          & 28\% \\
    \bottomrule
  \end{tabular}
  \caption{Distribution of quotation styles across literary works.}
  \label{tab:quotation_style_dist}
\end{table}

\subsection{Tunning Free Parameters}

As described in Section~\ref{Quotation Inference Stage}, ACT infers a quotation when its boosted score exceeds a predefined threshold. Setting this threshold too low increases recall but also results in many false positives—fragments misclassified as quotations. Conversely, a high threshold may reduce false positives but risk excluding genuine, especially shorter, quotations.

\begin{figure}[H]
\centering
\includegraphics[width=\linewidth]{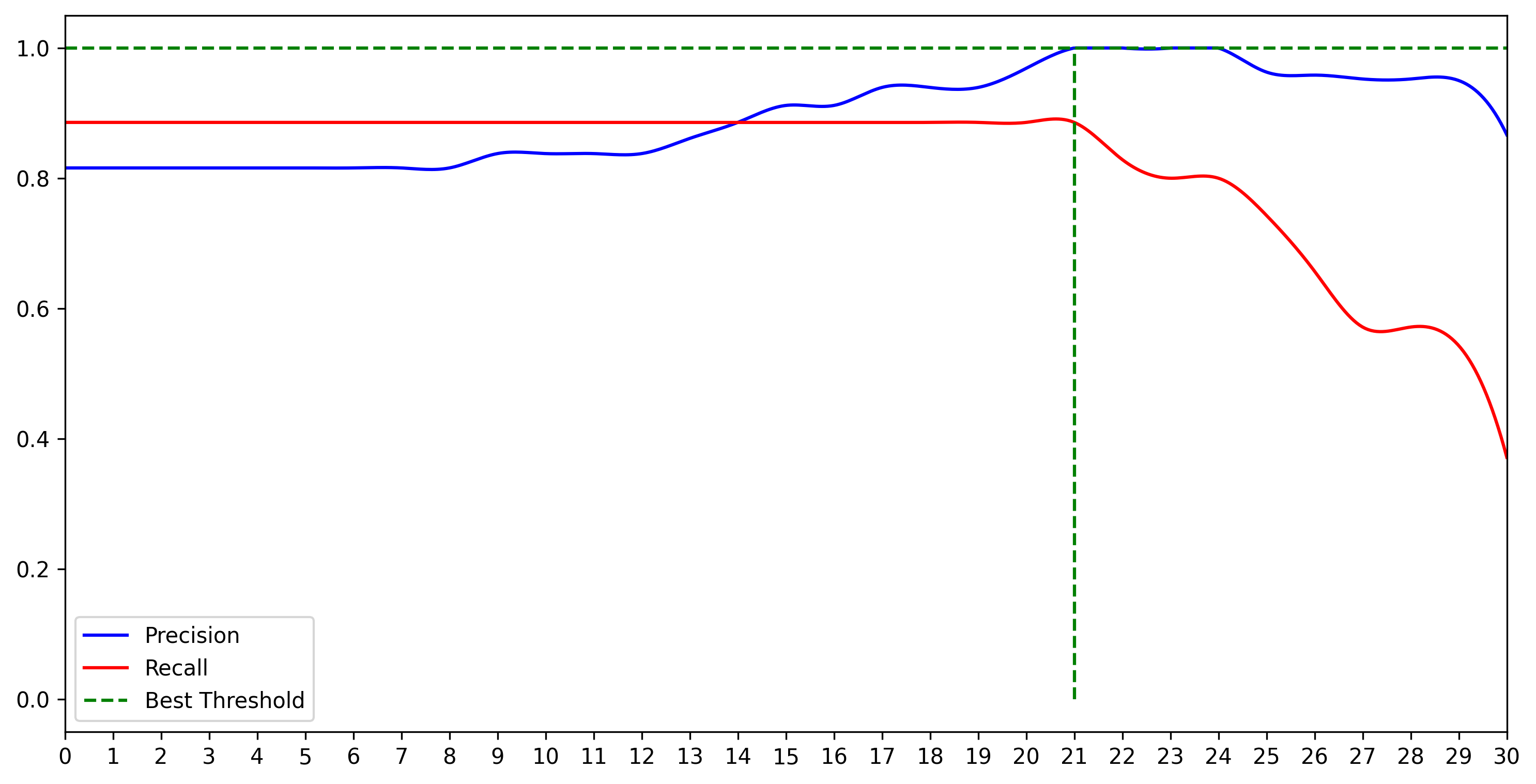}
\caption{Performance of ACT at various threshold values for quotation inference.}
\label{fig:Differenthresholds}
\end{figure}

To determine the optimal threshold, we performed a grid search on a held-out subset of the dataset. We varied the threshold and recorded precision and recall for each setting. Figure~\ref{fig:Differenthresholds} shows the results of this analysis. We observed that the optimal balance between precision and recall occurred when quotations with a score below 21 were discarded. This value was used as the default threshold in all subsequent experiments.

\section{Results}
\label{Results}

Table~\ref{tab:method_comp} presents the results of biblical quotation detection for various methods, including Precision, Recall, and $F_1$ score. The full version of our algorithm—ACT-QE—achieved the highest $F_1$ score (0.91), Recall (0.89), and Precision (0.94), clearly outperforming all baseline methods. 

\begin{table}[H]
    \centering 
        \begin{tabular}{lccl} 
            \toprule 
            Methods & $F_1$ & Recall & Precision \\
            \midrule 
            ACT-QE & \textbf{0.91} & 0.89 & \textbf{0.94} \\
            ACT-3 & 0.79 & 0.74 & 0.86 \\
            Dicta & 0.78 & 0.69 & 0.89 \\
            Passim & 0.62 & 0.55 & 0.71 \\
            ACT-2 & 0.59 & \textbf{0.90} & 0.44 \\
            Text-Matcher & 0.51 & 0.44 & 0.59 \\
            \midrule
            Manual Edition & 0.64 & 0.48 & 1.00 \\
            \bottomrule 
        \end{tabular} 
        \caption{Summary of biblical quotation detection results for the evaluated methods.} 
        \label{tab:method_comp} 
\end{table}

\subsection{ACT Configuration Analysis}
The performance of the different ACT configurations reveals the contribution of each innovation:

\begin{itemize}
    \item \textbf{ACT-2}, which includes only the morphology-aware alignment (Improvement A) with an n-gram size of 2, achieved the highest Recall ($0.90$) but a relatively low Precision ($0.44$). This configuration tends to over-identify quotations due to the absence of style-based filtering. Its high recall reflects its capacity to retrieve even short and morphologically altered quotations, but at the cost of many false positives.
    
    \item \textbf{ACT-3}, also limited to Improvement A but using an n-gram size of 3, showed more balanced performance with higher Precision ($0.86$) and a moderate Recall ($0.74$). Increasing the n-gram size reduces noise but limits the model's ability to detect shorter quotations, especially single-word quotations common in rabbinic texts.
    
    \item \textbf{ACT-QE}, which incorporates both innovations (A and B) with n-gram size 2, yielded the best balance across all metrics. The third stage—Quotation Enrichment—boosts quotations that exhibit structural patterns such as Wave and Echo, thereby improving both precision and recall. This demonstrates the value of combining syntactic alignment with contextual and structural inference.
\end{itemize}

These results validate our design decisions: the alignment algorithm enables robust matching in morphologically rich languages, and the additional quotation inference stage (Improvement B) is essential for refining results and reducing false positives.

\subsection{Comparison with Baseline Methods}
The critical editions achieved a Recall of only $0.48$, reflecting editorial priorities rather than exhaustive quotation annotation. Their perfect Precision ($1.00$) is a result of manual curation, in which only the most clearly defined quotations are included.

ACT outperforms Dicta, the leading baseline, by 12 percentage points in $F_1$ score. While Dicta achieves solid Precision ($0.89$), its Recall ($0.69$) is significantly lower than ACT’s. This indicates that Dicta misses many short or paraphrased quotations, which ACT successfully identifies.

Passim and Text-Matcher show lower performance ($F_1 = 0.62$ and $0.51$, respectively). Their design limitations—such as reliance on fixed n-gram schemes and lack of morphological awareness—hinder their ability to detect nuanced or non-verbatim quotations.

\subsection{Interpretation and Implications}
The high Recall achieved by ACT-QE demonstrates its effectiveness in identifying a broad range of relevant quotations, including short, paraphrased, or structurally embedded instances that are often missed by baseline systems. This capability is particularly important in the context of Midrashic literature, where quotations are frequently fragmented or reworded to serve sermonic or interpretive purposes. 

The comparative performance of ACT-2, ACT-3, and ACT-QE highlights that morphology-aware alignment alone is not sufficient for high-precision detection. The quotation enrichment stage introduced in ACT-QE plays a critical role in filtering false positives and accurately identifying complex stylistic structures.

\section{Conclusion}

This study introduced a novel pipeline for the automatic detection of biblical quotations in Hebrew texts, specifically targeting the nuanced and complex citation styles characteristic of Rabbinic literature. By overcoming the key limitations of existing frameworks—such as their inability to detect single-word quotations and reduced accuracy in fragmented or re-quoted passages—our method demonstrated substantial performance gains. The ACT algorithm achieved an $F_1$ score of 0.93, outperforming both the state-of-the-art Dicta system and critical editions by 11\% and 22\%, respectively.

A key innovation of our approach lies in its neighboring boosted scoring system, which has proven particularly effective for challenging styles, such as “Wave” and “Echo” quotations. This advancement not only improves detection precision but also offers new avenues for analyzing the stylistic and rhetorical structure of Rabbinic texts.

Beyond enhancing quotation detection, our findings suggest that the distribution of quotation styles may serve as a useful signal for literary classification. For example, the dominance of the “wave” style in \textit{Bereshit Rabba} and the prevalence of the “echo” style in \textit{Vayikra Rabba} point to stylistic patterns that could inform genre analysis and intertextual studies.

Furthermore, the methodological contrast between ACT and critical editions underscores a fundamental difference in aims: ACT strives for exhaustive, similarity-based detection, whereas human editors apply selective and interpretive judgment. The differences in annotation standards highlight the need to contextualize false positives in automated systems not merely as errors but as alternative forms of valid detection.

However, this study has several limitations that may affect the generalizability of its findings. The ground truth dataset, while carefully annotated by experts, was limited in scope, focusing on a select set of Midrashic compositions. As a result, further evaluation is needed to assess the robustness of ACT across a broader range of texts and historical contexts. Additionally, the classification of quotation styles into four categories—Simple, Wave, Echo, and Compound—is based on empirical patterns observed in our chosen corpora. Although these styles are well-represented in Rabbinic literature, they may not capture the full spectrum of quotation practices found in other textual traditions or genres. 

Future research should explore the applicability of the proposed quotation style categories in new domains and consider extending the typology to encompass additional forms of textual reuse. This includes evaluating ACT’s performance across different genres and languages, as well as analyzing its false positives in greater depth—distinguishing between lexical and conceptual paraphrases or stylistic transformations. Moreover, assessing human inter-annotator agreement could shed light on the boundary between interpretive discretion and algorithmic comprehensiveness, offering further insight into the nuanced challenges of automated quotation detection.

Looking forward, we aim to extend this framework to other languages and literary traditions, and to explore applications in stylometry and genre classification. Incorporating advanced natural language processing techniques—such as deep contextual embeddings—may further enhance the system’s robustness and sensitivity to subtle textual features.

Ultimately, this research contributes meaningfully to the automatic analysis of intertextuality in historical and religious texts, helping bridge the gap between traditional philological scholarship and modern computational methods. It lays the groundwork for a richer understanding of the literary, rhetorical, and cultural dimensions of Rabbinic literature within the broader field of digital humanities.

\section*{Data availability statement}
The data supporting the findings of this study are openly available in Zenodo at https://zenodo.org/record/15831340, reference number 15831340.

\section*{Author contributions}

Hadar Miller: Conceptualization, Methodology, Software, Formal analysis, Investigation, Resources, Data curation, Writing – Original Draft, Writing – Review \& Editing, Visualization, Supervision, Project administration.  
Tsvi Kuflik: Conceptualization, Methodology, Formal analysis, Writing – Review \& Editing.  
Moshe Lavee: Conceptualization, Methodology, Validation, Data curation, Writing – Review \& Editing.

\section*{Disclosure of interest}

The authors report there are no competing interests to declare.

\section*{Funding}

This research received no specific grant from any funding agency in the public, commercial, or not-for-profit sectors.

\bibliographystyle{tfnlm}
\bibliography{act}

\end{document}